\documentclass[letterpaper, 10 pt, one column,onference]{ieeeconf}  
\IEEEoverridecommandlockouts                              
\usepackage{graphicx}
\pdfoutput=1
\graphicspath{ {images/} }
\usepackage{subcaption}
\usepackage{amsmath}
\usepackage{amssymb}
\usepackage{float}
\usepackage{url}
\usepackage{lipsum}
\usepackage{algpseudocode}
\usepackage{algorithm}
\usepackage{xcolor}
\usepackage[colorinlistoftodos]{todonotes}
\usepackage[normalem]{ulem} 
\usepackage{cleveref}
\usepackage{cite}
\crefformat{section}{\S#2#1#3} 
\crefformat{subsection}{\S#2#1#3}
\crefformat{subsubsection}{\S#2#1#3}

\usepackage{paralist}
\usepackage[labelfont=bf,labelsep=space]{caption}

\definecolor{Junaed_color}{RGB}{255,0,0} 

\title{\LARGE \bf
Visual Diver Recognition for Underwater Human-Robot Collaboration
}

\author{Youya Xia$^{1}$ and Junaed Sattar$^{2}$
\thanks{The authors are with the Department of Computer Science and Engineering,
        The University of Minnesota, Minneapolis, MN, USA.
        {\tt\small \{$^{1}$xiaxxx244, $^{2}$junaed\} at umn.edu}}%
}

\begin{document}
\maketitle
\thispagestyle{empty}
\pagestyle{empty}
\begin{abstract}

This paper presents an approach for autonomous underwater robots to visually detect and identify divers. The proposed approach enables an autonomous underwater robot to detect multiple divers in a visual scene and  distinguish between them. Such methods are useful for robots to identify a human leader, for example, in multi-human/robot teams where only designated individuals are allowed to command or lean a team of robots. Initial diver identification is performed using the Faster R-CNN algorithm with a region proposal network which produces bounding boxes around the divers' locations. Subsequently, a suite of spatial and frequency domain descriptors are extracted from the bounding boxes to create a feature vector. A K-Means clustering algorithm, with $k$ set to the number of detected bounding boxes, thereafter identifies the detected divers based on these feature vectors. We evaluate the performance of the proposed approach on video footage of divers swimming in front of a mobile robot and demonstrate its accuracy.

\end{abstract}

\section{Introduction}
\label{sec:introduction} 

Underwater robotics is a rapidly expanding area of study in the field of autonomous robotics. Underwater robots are frequently used in a range of applications, including exploration, surveillance, and inspection tasks. However, due to the challenges and risks involved in the underwater domain and the current state of autonomous behaviors, remotely operated vehicles (ROVs) are most commonly deployed. Some autonomous underwater vehicles (AUVs) have also been used, \textit{e.g.}, for eliminating invasive species~\cite{dayoub2015robotic}. While ROVs provide a range of benefits, they require an operator on the `top-side' (on the surface of the body of water) to continuously operate the vehicle. The top-side operator is required to both interpret instructions coming from the divers and forward those instructions to the robot. This complicates the operational loop; adds significant temporal, monetary, and energy costs; and reduces the range of possible collaborative tasks.

Motivated by the desire to avoid such complex interaction methods, the authors' previous work has looked into protocols for direct human-robot interaction between divers and AUVs (\textit{e.g.},~\cite{Sattar08ICRA,Sattar14ICRA,islam2017mixed,islam2018Hands}) without the need for a top-side operator. Such protocols require methods for divers to communicate \textit{explicitly} with robots (for example, via hand gestures~\cite{islam2018Hands}), and also requires robots to \textit{implicitly} interact with divers by accompanying them during the missions~\cite{Sattar09RSS,islam2017mixed}. 

\begin{figure}[thp]
	\vspace{2.5mm}
	\centering
	\begin{subfigure}[b]{0.49\linewidth}
		\includegraphics[width=\linewidth]{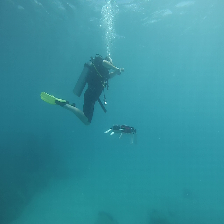}
		\phantomcaption
		\label{fig:js_bbd2018}	
	\end{subfigure}
	\begin{subfigure}[b]{0.49\linewidth}
		\includegraphics[width=\linewidth]{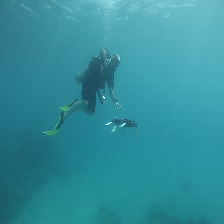}
		\phantomcaption
		\label{fig:chris_bbd2008}
	\end{subfigure}
    \vspace{-5mm}
	\caption{A sequence of images showing a diver and robot collaborating directly. In such missions, an AUV often needs to not just follow \textit{any} diver but a \textit{specific} diver.}
	\label{fig:js_allrobots}
\end{figure}

Detecting a diver or swimmer in underwater environments poses a significant challenge to vision-based methods due to optical distortions, color absorption, and scattering issues. Sensors relying on electromagnetic emissions (\textit{e.g.}, radar, lidars, radio, wifi) are susceptible to large attenuation and are thus unusable for underwater applications. Sonar is predominantly used in many underwater vehicles, particularly for localization, long-range sensing, and low-bandwidth communication, but does not provide the bandwidth and richness necessary for AUV's to track targets in real-time. Furthermore, active sensors can be intrusive to marine species and have detrimental effects on their well-being. With recent advances in deep machine learning, particularly in convolutional and recurrent neural networks, generative adversarial networks, and deep reinforcement learning, recent development in machine vision have shown some promising results in underwater applications. In particular, robot convoying~\cite{shkurti2017underwater}, image enhancement~\cite{Fabbri2018ICRA}, and gesture-based programming~\cite{islam2018Hands} have been shown to work well in real-world settings. High accuracy with deep object detectors (\textit{e.g.},~\cite{redmon2016yolo9000,NIPS2015_5638}), and the availability of embedded, power-efficient hardware that can efficiently run with deep models have encouraged robotics researchers to delve into the `tracking-by-detection' approach. However, while these methods are able to robustly detect objects of interest in a scene (divers in our case), they have not been able to distinguish between them unless there is a high degree of `in-class' feature  diversity. In other words, individual detected objects, while belonging to the same class, should exhibit difference in features to be distinguished robustly. In the case of divers in underwater scenes, such feature diversity is often absent. In addition, data scarcity is an issue that prevents deep methods from reliably identifying individual divers. Due to the very nature of deep learning methods, little control can be asserted over the feature selection process, which makes them a somewhat less desirable choice.

This paper presents a method that not only visually tracks swimmers and divers but is able to uniquely identify them. A convolutional model-based object detector, specifically Faster R-CNN with region proposal network, is first used to detect divers in the scene, and bounding boxes in the image containing the divers are generated. These bounding boxes are subsequently passed on to a suite of feature detectors comprised of spatial and frequency-domain image features. The vectors constructed by the said detectors are then fed into a K-nearest neighbor clustering algorithm to identify individual divers. Specifically, this work contributes the following:
\begin{enumerate}
	\item a method for visually detecting and identifying divers underwater;
	\item a method combining supervised feature-based and feature learning with unsupervised learning for diver identification;
	\item a real-time implementation of the said algorithm to run on-board a mobile robot\footnote{{\small\url{https://github.com/xiaxx244/diver_detection.git}}}; and
	\item extensive evaluation of the method on datasets of divers and swimmers collected from a variety of locations and environmental conditions.
\end{enumerate}
The task of identifying individual divers, as stated previously, is both open and challenging, and is required for human-robot collaborative tasks underwater. The proposed work is the first of its kind to achieve this by learning diver features from visual stimuli using both deep and feature-engineered methods. 
\vspace{2mm}
\section{Related Work}
\label{sec:related}
This work is a combination of people tracking and identification tasks; a rich body of literature exists in this domain~\cite{islam2018person}. Niyogi and Adelson~\cite{niyogi_analyzing_1994} use the positions of the head and ankles to detect human walking patterns orthogonal to camera view direction. In the seminal work using ``moving light displays'', Rashid observed~\cite{Rashid1980} that human visual systems are quite sensitive to even limited human-like motions. Identifying walking gaits have also been investigated, as shown in recent advancements in Biometrics~\cite{nixon_human_2005}. Automated analysis of walking gaits~\cite{Sidenbladh00Stochastic,Sidenbladh2003} have also yielded promising results.


The Kalman filter~\cite{Kalman60} is the classical approach for real-time tracking. However, a linear dynamics model of the given system is required for it to work. The motion of human swimmers is quite non-linear and linearization of the system model may lead to subpar performance or in the worst case divergence. The Unscented (otherwise known as the Sigma-Point) Kalman Filter~\cite{julier97new} allows for some non-linearity in the tracked process and is less computationally expensive than fully non-parametric algorithms (\textit{e.g.},~\cite{Isard98Condensation}).


Visual tracking of divers and swimmers has not been explored greatly though work exists for visual tracking of arbitrary targets and subsequent robot servoing~\cite{hutchinson96tutorial}. Also, real-time control and tracking schemes have been shown to work well for visual target-following underwater (\textit{e.g.}, ~\cite{Sattar05IROS}). Spatio-temporal tracking of biological motion for diver-following has been shown to work when divers swim directly away from the robot~\cite{Sattar07IROS} and in other straight-line trajectories~\cite{Sattar09RSS}. Recent work has also made it possible to track divers swimming in arbitrary directions~\cite{islam2017mixed}.

Deep visual models for target detection have seen rapid adoption of late and have shown high-accuracy in a number of challenging tasks. In this work, we use Faster R-CNN~\cite{NIPS2015_5638} with a region proposal network for finding diver locations. However, researcher have developed a number of other accurate models such as the Mask R-CNN~\cite{he2018mask}, Single Shot MultiBox Detector (SSD)~\cite{liu2016ssd}, and a family of You Only Look Once (YOLO) models (YOLO V2~\cite{redmon2016yolo9000}, Tiny YOLO \cite{tinyYOLO}, etc.). These are the fastest (in terms of processing time of a single frame) among the family of current state-of-the-art models~\cite{tfzoo} for general object detection. We train these models using a rigorously prepared dataset containing sufficient training instances to capture variations of diver appearances that can arise in underwater human-robot collaborative scenarios.

\section{Methodolgy}
\label{sec:methodology} 

The proposed algorithm for identifying divers is detailed in the following subsections. In particular, we explain the feature-based unsupervised identification process of divers and the factors that lead to those choices in detail.

\subsection{Diver Detection using Deep Models}
\label{sec:deep_detection_model}
In order to construct a feature vector to distinguish each diver, we need to find all divers inside an image. Typically, the methods which can be used to find pedestrians or people in terrestrial scenes tend to fail when trying to detect a diver since the shape of a diver is different from the shape of a pedestrian. This difficulty arises from posture differences as divers are in predominantly horizontal orientations underwater. The additional gear worn by the divers (\textit{e.g.}, dive suits, buoyancy devices, fins) also creates challenges for such algorithms. The authors' previous work has looked at periodic motion cues for diver detection (\textit{e.g.}, the Mixed Domain Periodic Motion or MDPM~\cite{islam2017mixed} algorithm) and it has been shown to work well. However, MDPM does not generate a bounding box around the diver, as it tracks the propagation of the energy signature in the frequency domain generated by the diver's swimming gait. Therefore, in order to detect a diver, instead of more traditional approaches (such as HOG (histogram-of-gradient) descriptors), we opted for a deep learning model to detect divers in a scene with bounded locations.

Using the principles of a convolutional neural network (CNN), an input neuron in Faster R-CNN is only connected to part of the first layer of the network. However, Faster R-CNN adds a region proposal network just before the object classifier CNN to generate anchor boxes (\textit{i.e.}, potential bounding boxes). Therefore, only such bounding boxes are needed to be given to a smaller (`shallower') CNN which is designed for classification and regression, making it faster than using a full CNN over the entire image space. These features, along with the accuracy shown by Faster R-CNN, made it a prime choice for the diver detection phase of the proposed algorithm. For the purpose of training our diver detection model, we used approximately $2000$ labeled images of divers in underwater settings. These images were obtained from field trials we conducted at previous times in pools, lakes, and oceans over the past few years. While $2000$ images may not seem sufficient to train a deep detection algorithm, having a pretrained Faster R-CNN model makes it possible to achieve high accuracy by simply using the additional training data for the required object class (divers in our case).
\begin{figure}[h]
	\centering
	\begin{subfigure}[b]{0.48\linewidth}
		\centering %
		\includegraphics[width=\linewidth]{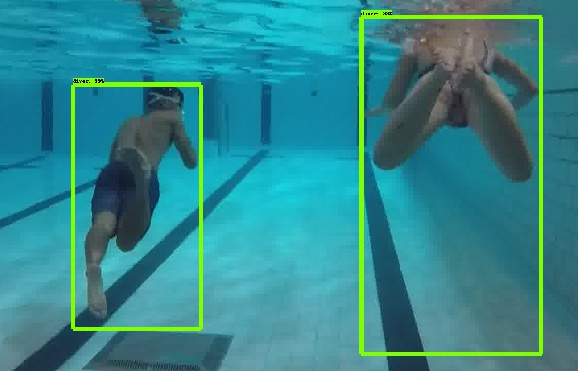}
	\end{subfigure}
	\begin{subfigure}[b]{0.48\linewidth}
		\centering	
		\includegraphics[width=\linewidth]{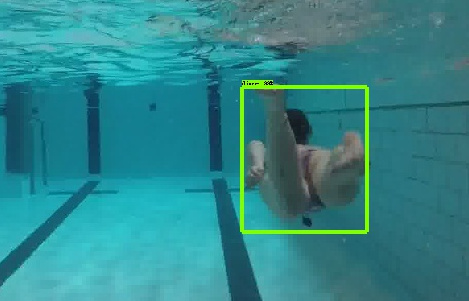}
	\end{subfigure}
	\caption{Bounding boxes around divers after detection.}
	\label{fig:SwimDetect}
\end{figure}
\vspace{-2mm}

\subsection{Feature Extraction}
Once divers are detected in an image, we need to construct a feature vector for each detected diver. Our approach here is to use feature-based unsupervised learning to classify each bounding box returned by Faster R-CNN to individually identify each diver. The following sections describe the features chosen for this purpose. 

\paragraph{Average Color Distribution}
While color as a standalone feature can be affected greatly by optical distortions and attenuation, it can be a useful discriminator within bounding boxes containing divers. Specifically, complexion and the colors of the dive suit and gear can be valuable cues. Although RGB values may be a good indicator for identifying the color differences between divers, we convert the color space from RGB to LAB to provide more precise color comparison. LAB is a three-dimensional color space which represents lightness of color, position between red and green, and position between yellow and blue. However, only using the sum of LAB values of each pixel inside each bounding box may lead to incorrect classification when a diver's distance from a robot changes. Therefore, instead of using the sum of LAB values, we choose to use the average LAB value inside each bounding box. If $\mu$ is the average color in each bounding box, $x$ and $y$ are the horizontal and vertical coordinates respectively of each box and $l$, $a$ and $b$ are the LAB values of each pixel, then 
\begin{equation} 
\mu = \frac{l+a+b}{(y_{max}-y_{min}) \times (x_{max}-x_{min})}
\end{equation} 

Additionally, to improve overall precision, each bounding box is divided into four equal rectangular regions and the average color values for each region are obtained separately. The final feature vector contains four average LAB values (\textit{i.e}., $\mu_{i}$, where $i=1\rightarrow 4$) as a color descriptor of the diver. 


\paragraph{Amplitude of Spatial Frequency Distribution}
We also look at spatial frequency of diver's features to extract unique signatures, by using the two-dimensional Fourier Transform. The two-dimensional Fourier Transform of an image can be formulated as:

\begin{equation}
F(k,l)=\sum_{i=0}^{N-1} \sum_{j=0}^{N-1}  f(i,j)e^{-i2 \pi \frac{(ki+lj)}{N}}
\end{equation}
where $f(i,j)$ is the image in the spatial domain and the exponential term is the base function corresponding to each point $F(k,l)$ in the Fourier space.

After applying the 2D Fast Fourier Transform, we compute the average amplitude of each diver and use the three average amplitudes (for each R, G, and B channel) as diver features.


\begin{figure}[b]
	\centering
	\begin{subfigure}[b]{0.2\textwidth}
		\centering	\vspace*{2mm}
		\includegraphics [width=37mm,height=29mm]{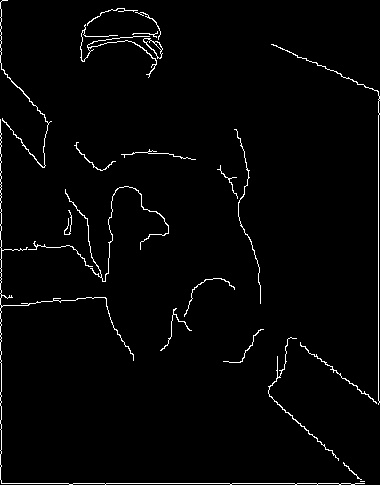}
	\end{subfigure}
	\hspace{2mm}
	\begin{subfigure}[b]{0.2\textwidth}
		\centering	
		\includegraphics[width=37mm,height=29mm]{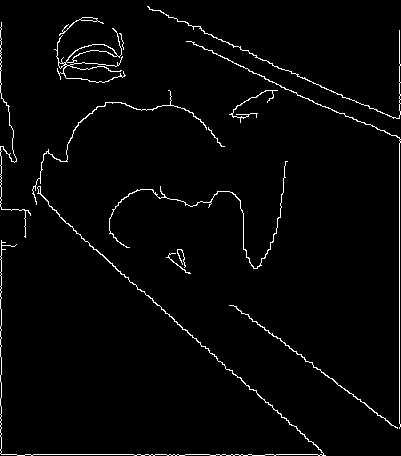}
	\end{subfigure}
	
	\begin{subfigure}[b]{0.2\textwidth}
		\centering	\vspace*{2mm}
		\includegraphics[width=37mm,height=29mm]{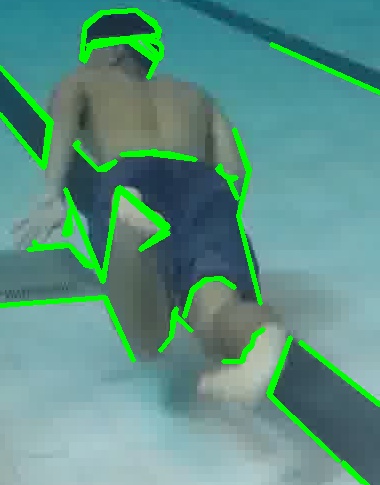}
	\end{subfigure}  
	\hspace{2mm}
	\begin{subfigure}[b]{0.2\textwidth}
		\includegraphics[width=37mm,height=29mm]{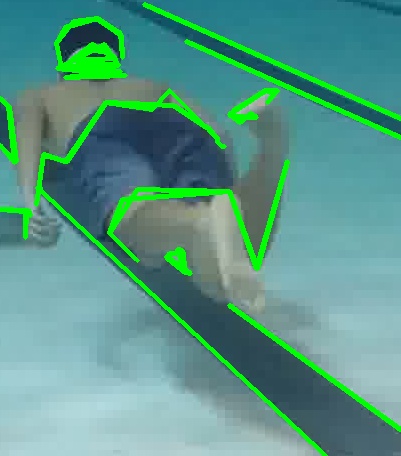}
	\end{subfigure} 
	\caption{Sequence of edge features (top) and contours within the bounding box for diver Liam.}
	\label{fig:cannyEdgeDetect1}	
\end{figure}

\begin{figure}[t]
	\centering
	\begin{subfigure}[b]{0.2\textwidth}
		\centering	\vspace*{2mm}
		\includegraphics [width=37mm,height=29mm]{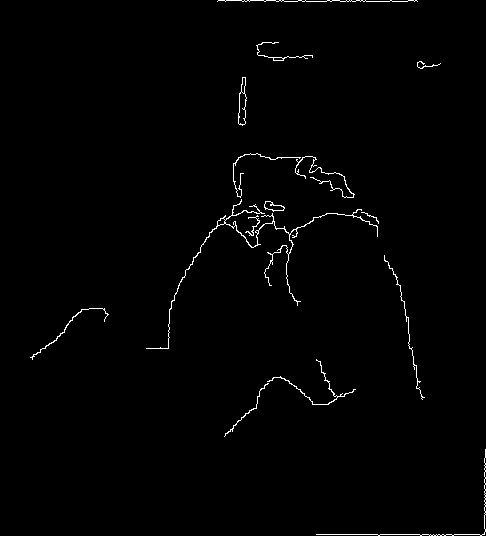}
	\end{subfigure}
	\hspace{2mm}
	\begin{subfigure}[b]{0.2\textwidth}
		\centering	
		\includegraphics [width=37mm,height=29mm]{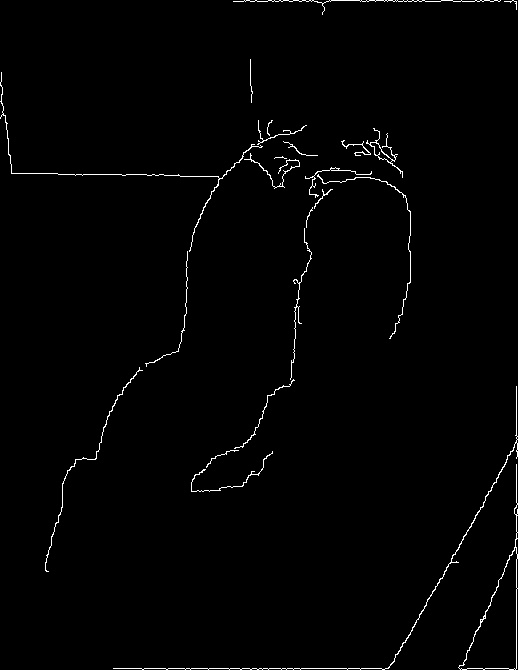}
	\end{subfigure}
	
	\begin{subfigure}[b]{0.2\textwidth}
		\centering	\vspace*{2mm}
		\includegraphics [width=37mm,height=29mm]{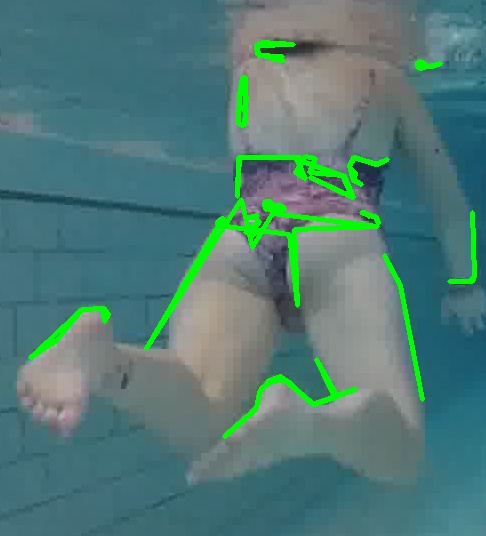}
	\end{subfigure}  
	\hspace{2mm}
	\begin{subfigure}[b]{0.2\textwidth}
		\centering	\vspace*{2mm}
		\includegraphics [width=37mm,height=29mm]{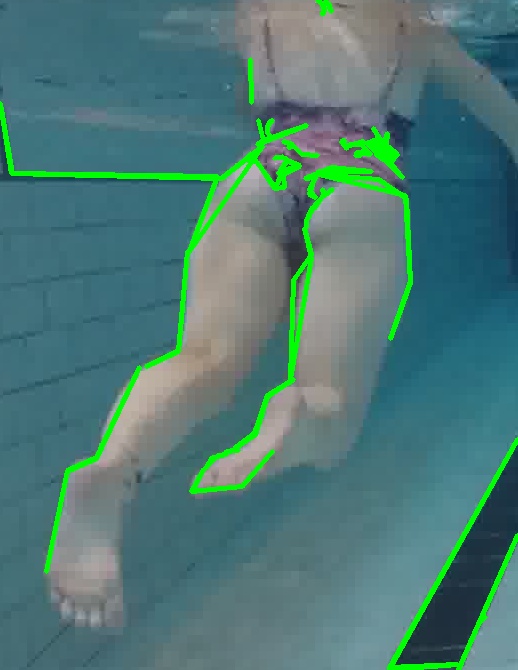}
	\end{subfigure} 
	\caption{Sequence of edge features (top) and contours within the bounding box for diver Emma.}
	\label{fig:cannyEdgeDetect2}
\end{figure}

\paragraph{Shape Approximation using Edge Features}
This feature aims to capture the differences in divers' physiques -- shape in particular -- factoring in the effect of the dive gear. In order to achieve this goal, we need to extract the diver's contours within the bounding box. The Canny edge detection algorithm\cite{bao2005canny} is first applied to extract the edges within the diver's bounding box after smoothing the area using a Gaussian filter. For each detected edge, the \textit{Ramer-Douglas-Peucker} (RDP) algorithm~\cite{douglas1973algorithms} is applied to approximate the edges with fewer points. Finally, the average value (which is a 2-tuple, \textless$\bar{E_x},\bar{E_y}$\textgreater) of all approximated points in all contours is used as a feature. The sequence of edge features of two divers and their corresponding contours shown in Figures~\ref{fig:cannyEdgeDetect1} and~\ref{fig:cannyEdgeDetect2} demonstrate the differences between the two divers. The pool markers do get included in the feature set; which may adversely affect detection accuracy. However, in most cases, pool markers do not add significantly to each swimmer's feature set, and their effect is further marginalized by computing the average of all the contour points (for both convex hulls and edge features). Therefore, even with those markers, the edge and convex hull features for each diver are quite unique and provide distinctive features.

\begin{figure}[t]
	\centering
	\begin{subfigure}[b]{0.2\textwidth}
		\centering	\vspace*{2mm}
		\includegraphics [width=37mm,height=29mm]{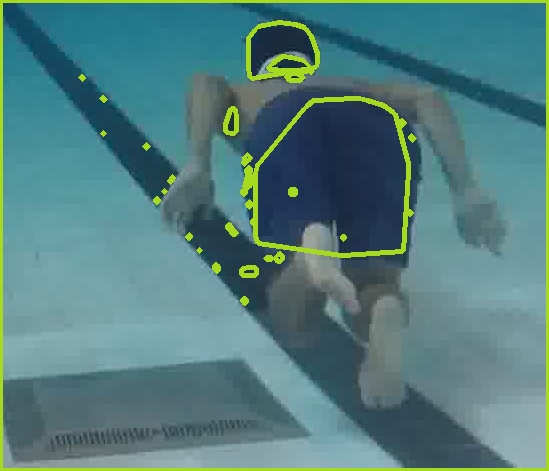}
	\end{subfigure}
	\hspace{2mm}
	\begin{subfigure}[b]{0.2\textwidth}
		\centering	
		\includegraphics [width=37mm,height=29mm]{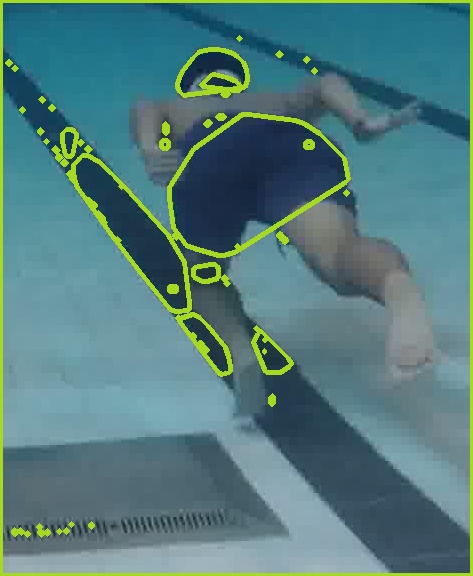}
	\end{subfigure}
	
	\begin{subfigure}[b]{0.2\textwidth}
		\centering	\vspace*{2mm}
		\includegraphics [width=37mm,height=29mm]{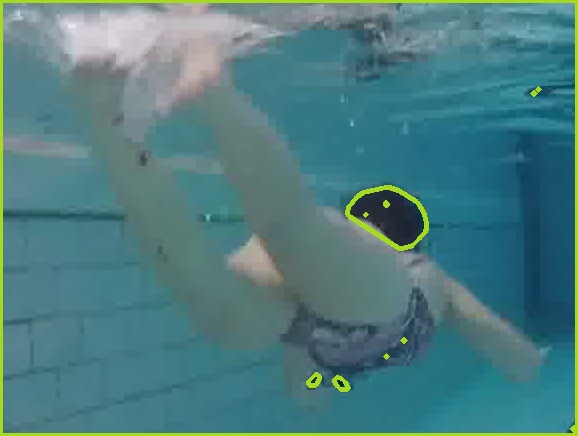}
	\end{subfigure}  
	\hspace{2mm}
	\begin{subfigure}[b]{0.2\textwidth}
		\centering	\vspace*{2mm}
		\includegraphics [width=37mm,height=29mm]{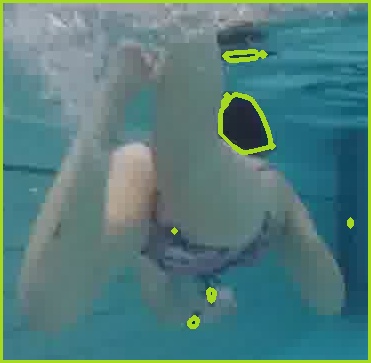}
	\end{subfigure} 
	\caption{Convex hull features on divers Liam (top row) and Emma (bottom row), drawn in yellow overlays.}
	\label{fig:convexHull}
	
\end{figure}

\paragraph{Shape Approximation using the Convex Hull}
A convex hull is defined as a convex polygon constructed by obtaining a minimal subset of the points such that all the points in the set fall either inside or on the boundary of the polygon~\cite{berg2008computational}. In order to obtain a convex hull, we first convert the bounding box image to grayscale and apply a threshold to suppress pixels which have significantly low intensity (specifically pixels having intensity values of $50$ or lower in a scale of $0$ to $255$). A subsequent step extracts contours from this \textit{binary} image in a compressed format, preserving contour hierarchies. We compute convex hulls of all these contours using the Gift Wrapping algorithm~\cite{cormen2009introduction}, using this compressed representation of contour points as input. The theory is that the  number and shapes of convex hulls drawn for each diver will capture the variability inherent in the shapes of divers. As in the edge features, the average of all points on each of these hulls, \textless$\bar{C_x}, \bar{C_y}$\textgreater is used as a feature. Figure~\ref{fig:convexHull} shows some results of convex hulls constructed on divers' outlines. Note that the convex hulls for each diver are significantly different and are dependent on their posture and physique. For instance, there are two major convex hulls drawn for Liam (one around the head and one on the bottom), whereas there is only one major convex hull drawn for Emma.

\paragraph{Image Moments}
In image processing, an image moment is defined as the weighted average of intensities in an image. Hu proposes seven specific moments which have been shown to be invariant to changes in translation, rotation, and scale~\cite{hu1962visual}. Since these Hu's moments will remain unchanged for a specific diver even if the the diver's orientation or the distance between the diver and the robot changes, they are strong candidates to be used as unique features of divers. These seven moments are computed for each diver's bounding box and used as feature descriptors.

We evaluated other feature descriptors (such as ORB~\cite{rublee2011orb} and SURF~\cite{bay2006surf}) but these failed to provide sufficient distinguishing ability and were not ultimately used.
 
\subsection{K-Means clustering}
Once diver bounding boxes are obtained and feature vectors consisting of the above-mentioned features are constructed, we use the K-Means clustering~\cite{berg2008computational} implemented by  Lloyd’s algorithm~\cite{lloyd1982least} to cluster all feature vectors obtained from diver regions. Note that in the K-Means clustering algorithm, the number of clusters $K$ needs to be chosen upfront. Since it is possible that the general diver detection of the initial frame may not identify all possible divers in the whole detection process (\textit{e.g.}, some divers may appear in the middle of the detection process or the general diver detection does not capture all divers in the initial frames), we decide to choose the number $K$ to be the maximum number of divers appearing during the detection process. During the tracking process, the initial cluster centers are randomly assigned to the collected feature vectors. During each subsequent iteration, K-Means assigns each feature vector to its closest cluster center using its Euclidean norm and recomputes each cluster center to be the mean among the feature vectors assigned to its group. The cluster refinement process stops after cluster centers converge with error falling below a threshold of $1e^{-4}$.
\section{Experiments}
\label{sec:experiments} 
We evaluate the performance of the proposed approach using video data of divers in different bodies of water and visual conditions, in both open-water (\textit{e.g.}, oceans, lakes) and closed-water (\textit{e.g.}, swimming pools) settings. In this section, we discuss the details of the validation process and the subsequent results.

\begin{table*}[thp]
	\begin{center}
		\begin{tabular}{ | p{9cm} | c | c | c | } 
			\hline
			Scenario & Accuracy(\%) & \parbox[t]{2cm}{Missed\\Identification(\%)} & \parbox[t]{2cm}{Wrong\\Identification(\%)} \\ 
			\hline \hline
			Scenario 1: two divers, no flippers, one diver exits scene & 100 & 0 &0 \\ 
			\hline
			Scenario 2: two divers, no flippers, one diver exits scene and later reenters & 96.8 & 0 &3.2 \\ 
			\hline
			Scenario 3: two divers, with flippers, one diver exits scene & 94.9 & 0.3 & 4.8 \\ 
			\hline
			Scenario 4: two divers, with flippers, one diver exits scene and later reenters & 90.8 & 2.2 & 7\\ 
			\hline
			Scenario 5: three divers, no flippers, one diver exits scene & 77.5 & 1.4 &21.1\\ 
			\hline
			Scenario 6: three divers, with flippers, one diver exits scene & 80.7 & 0 &19.3 \\ 
			\hline
			Scenario 7: two divers, no flippers, freeform swim & 90.5 & 0 & 9.5 \\ 
			\hline
			Scenario 8: two divers, ocean waters, full-body dive suit and flippers & 96.07 & 0 &3.93 \\ 
			\hline
		\end{tabular}
		\caption{Quantitative performance of the proposed diver identification algorithm in different environmental conditions with a varying number of divers.}
		\label{table:performanceTable}
	\end{center}
\end{table*}

\subsection{Experimental Setup}
In order to evaluate the performance of the proposed approach, we conducted several pool trials with a varying number of people in the scene. Images were captured using handheld underwater cameras (\textit{e.g.}, GoPros\texttrademark) or a trailing underwater robot. 

The experiments were set up in two different scenarios. In the first case, two divers are seen swimming together \textit{without flippers} at the beginning of the experiment. About halfway through the sequence, one diver leaves the scene and does not return, leaving the other one swimming solo until the end of the sequence. The second scenario begins similarly, with two divers swimming together. About a third into the sequence, one of the divers leaves the scene, while the remaining diver continues swimming. However, unlike the first scenario, the second diver reappears in the scene about two-thirds into the sequence and continues swimming together with the first diver until the end of the sequence. The two scenarios were repeated in another trial where both divers wore flippers to evaluate the algorithm's performance under subtle diver appearance changes.  

We also conduct the entire experiment as described above with three divers instead of two, having one diver leave and reappear as before, leaving two divers consistently swimming throughout.

\subsection{Deep Diver Detection Model}
\begin{figure}[b]
	\centering
	\begin{subfigure}[b]{0.2\textwidth}
		\centering	\vspace*{2mm}
		\includegraphics [width=37mm,height=29mm]{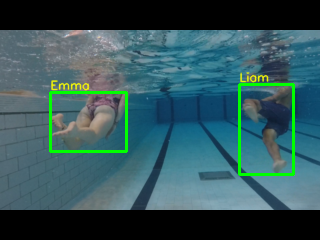}
	\end{subfigure}
	\hspace{2mm}
	\begin{subfigure}[b]{0.2\textwidth}
		\centering	
		\includegraphics [width=37mm,height=29mm]{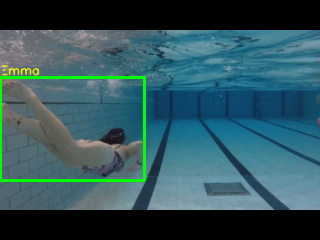}
	\end{subfigure}
	\begin{subfigure}[b]{0.25\textwidth}
		\centering	\vspace*{2mm}
		\includegraphics [width=37mm,height=29mm]{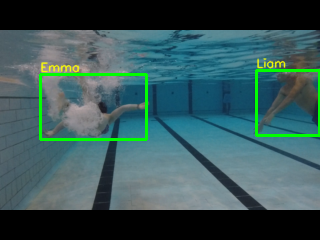}
	\end{subfigure}  
	\hspace{2mm}
	\begin{subfigure}[b]{0.2\textwidth}
		\includegraphics [width=37mm,height=29mm]{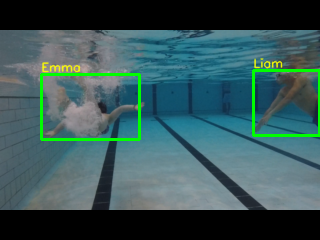}
	\end{subfigure} 
	\caption{Divers without flippers. Top row: Liam and Emma are both detected, and then Liam leaves the scene. Bottom
		row: Liam correctly identified after he reappears.}
	\label{fig:twoDiversNoFlippersResume}
\end{figure}

During the general diver detection stage, we used the Faster R-CNN~\cite{NIPS2015_5638} model with pretraining using a join-training scheme~\cite{chen2017implementation}, which requires less additional training data. As mentioned in Section~\ref{sec:deep_detection_model}, we use $2000$ labeled images of divers for training the general diver detection model. Images from the datasets collected during the pool trials were used for testing. In addition, to compare the performance of our algorithm under different visual conditions, we used datasets collected from previous pool and ocean trials conducted at the Bellairs Research Center in Barbados\footnote{{\small \url{https://www.mcgill.ca/bellairs/}}}. The Faster R-CNN model has been observed to work well, achieving about 98\% accuracy in our test datasets. Figure~\ref{fig:SwimDetect} shows the output of the deep detection model. The bounding boxes shown in Figures~\ref{fig:twoDiversNoFlippersResume} to~\ref{fig:twoDiversOceanTwo} are also detected using the same method, which demonstrate its effectiveness in different environmental conditions. 

\subsection{Diver Feature Selection}
We have visually demonstrated some of the features used for diver identification in Figures~\ref{fig:cannyEdgeDetect1} to~\ref{fig:convexHull}. In the subsequent discussion, we arbitrarily name the divers as Emma, Noah, and Liam. The goal is to consistently identify each diver with the same label each time they are visible in the scene. Using a reliable object detector like Faster R-CNN ensures that the location of the diver can be accurately found (seen in Figure~\ref{fig:SwimDetect}). This in turn assists with the construction of the diver's feature vector and subsequent diver identification. 

\subsection{Recognition Accuracy}
\begin{figure}[t]
	\centering
	\begin{subfigure}[b]{0.2\textwidth}
		\includegraphics [width=37mm,height=29mm]{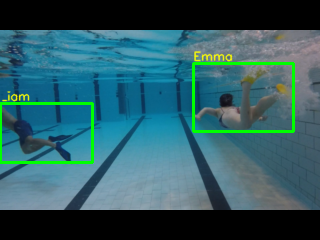}
	\end{subfigure}
	\hspace{2mm}
	\begin{subfigure}[b]{0.2\textwidth}
		\centering	
		\includegraphics [width=37mm,height=29mm]{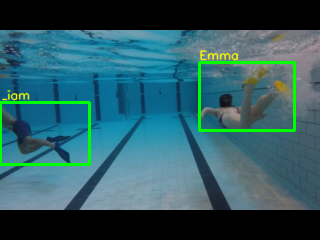}
	\end{subfigure}
	\begin{subfigure}[b]{0.25\textwidth}
		\centering	\vspace*{2mm}
		\includegraphics [width=37mm,height=29mm]{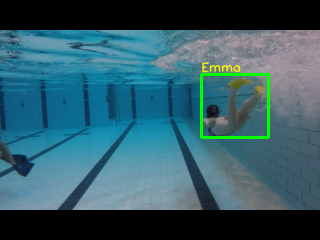}
	\end{subfigure} 
	\hspace{2mm}
	\begin{subfigure}[b]{0.2\textwidth}
		\centering	\vspace*{2mm}
		\includegraphics [width=37mm,height=29mm]{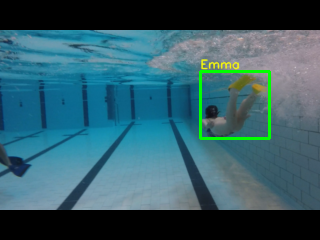}
	\end{subfigure} 
	\caption{Divers with flippers. Top row: Liam and Emma are both detected. Bottom row: Emma correctly identified after Liam leaves the scene.}
	\label{fig:twoDiversFlippersNoResume}
\end{figure}

\begin{figure}[t]
	\centering
	\begin{subfigure}[b]{0.2\textwidth}
		\includegraphics [width=37mm,height=29mm]{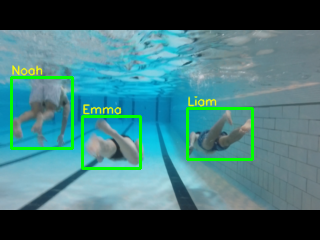}
	\end{subfigure}
	\hspace{2mm}
	\begin{subfigure}[b]{0.2\textwidth}
		\centering	
		\includegraphics [width=37mm,height=29mm]{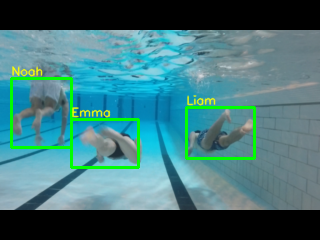}
	\end{subfigure}
	\begin{subfigure}[b]{0.25\textwidth}
		\centering	\vspace*{2mm}
		\includegraphics [width=37mm,height=29mm]{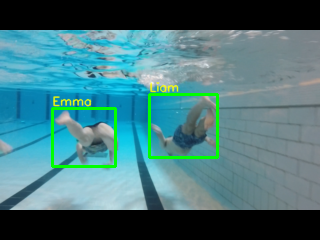}
	\end{subfigure}  
	\hspace{2mm}
	\begin{subfigure}[b]{0.2\textwidth}
		\centering	\vspace*{2mm}
		\includegraphics [width=37mm,height=29mm]{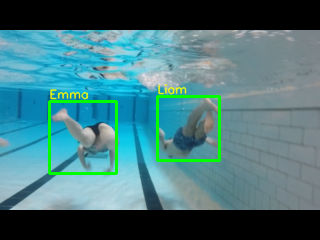}
	\end{subfigure} 
	\caption{Divers without flippers. Top row: Liam, Noah and Emma are all detected, and then Liam leaves the scene. Bottom row: Noah and Emma correctly identified.}
	\label{fig:threeDiversNoFlippersNoResume}
\end{figure}

\begin{figure}[h]
	\centering
	\begin{subfigure}[b]{0.2\textwidth}
		\includegraphics [width=37mm,height=29mm]{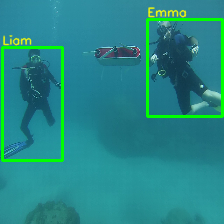}
	\end{subfigure}
	\hspace{2mm}
	\begin{subfigure}[b]{0.25\textwidth}
		\centering	
		\includegraphics [width=37mm,height=29mm]{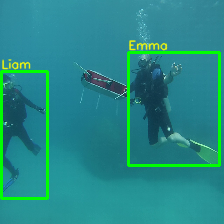}
	\end{subfigure}
	\begin{subfigure}[b]{0.2\textwidth}
		\centering	\vspace*{2mm}
		\includegraphics [width=37mm,height=29mm]{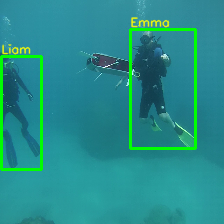}
	\end{subfigure}  
	\hspace{2mm}
	\begin{subfigure}[b]{0.2\textwidth}
		\centering	\vspace*{2mm}
		\includegraphics [width=37mm,height=29mm]{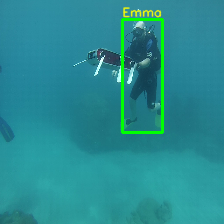}
	\end{subfigure} 
	\caption{Divers Emma and Liam in SCUBA gear in the ocean; Liam gradually disappears from the scene without affecting detection accuracy.}
	\label{fig:twoDiversOceanNoResume}
\end{figure}

\begin{figure}[h]
	\centering
	\begin{subfigure}[b]{0.2\textwidth}
		\includegraphics [width=37mm,height=29mm]{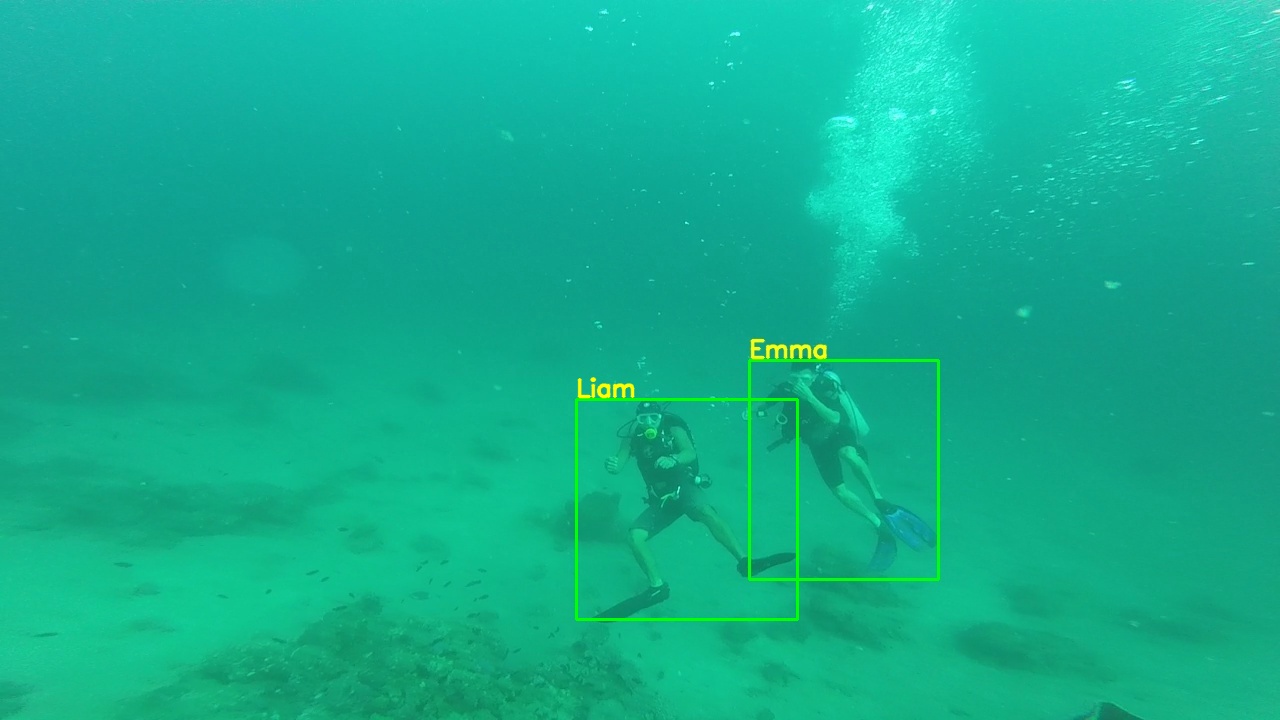}
	\end{subfigure}
	\hspace{2mm}
	\begin{subfigure}[b]{0.2\textwidth}
		\centering	
		\includegraphics [width=37mm,height=29mm]{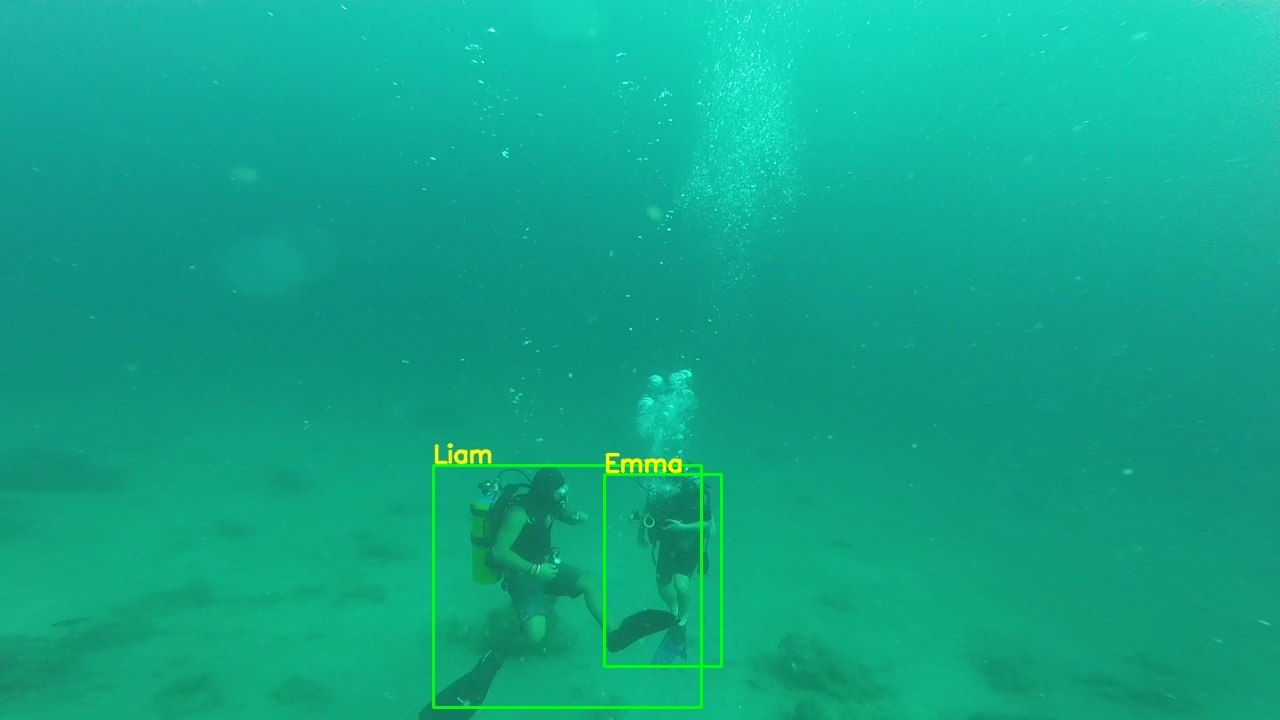}
	\end{subfigure}
	\begin{subfigure}[b]{0.25\textwidth}
		\centering	\vspace*{2mm}
		\includegraphics [width=37mm,height=29mm]{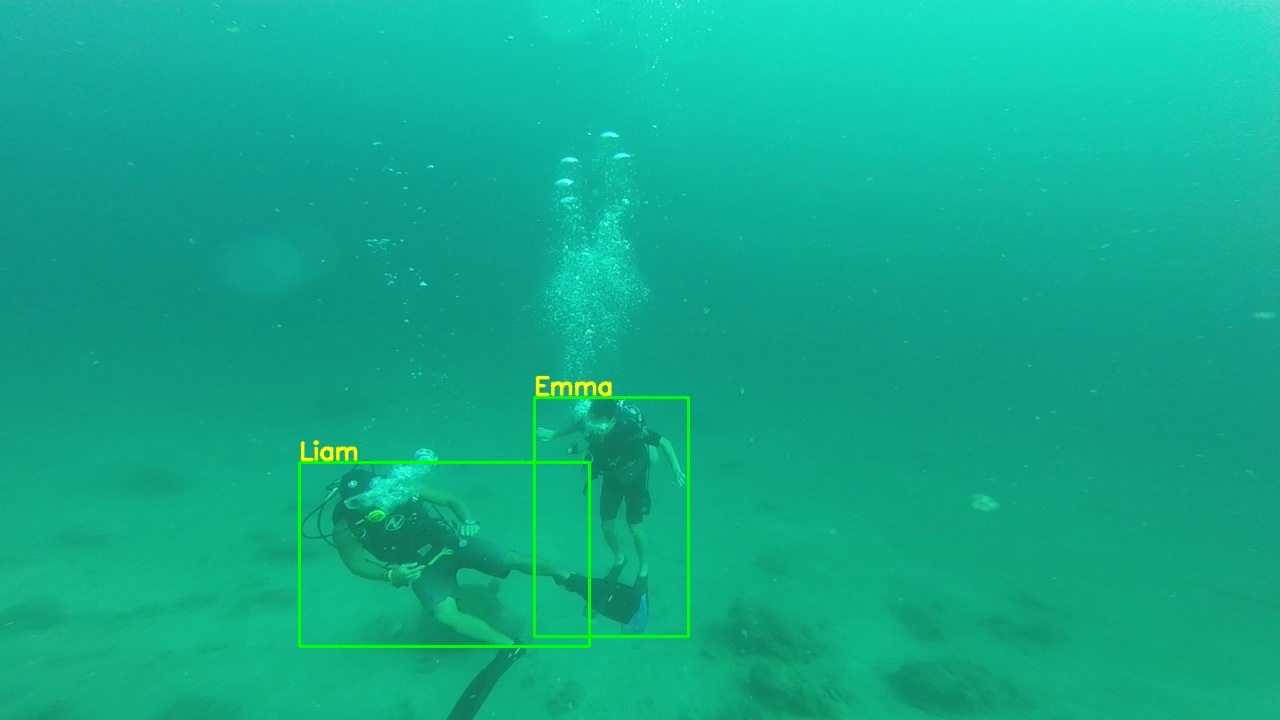}
	\end{subfigure}
	\hspace{2mm}
	\begin{subfigure}[b]{0.2\textwidth}
		\centering	\vspace*{2mm}
		\includegraphics [width=37mm,height=29mm]{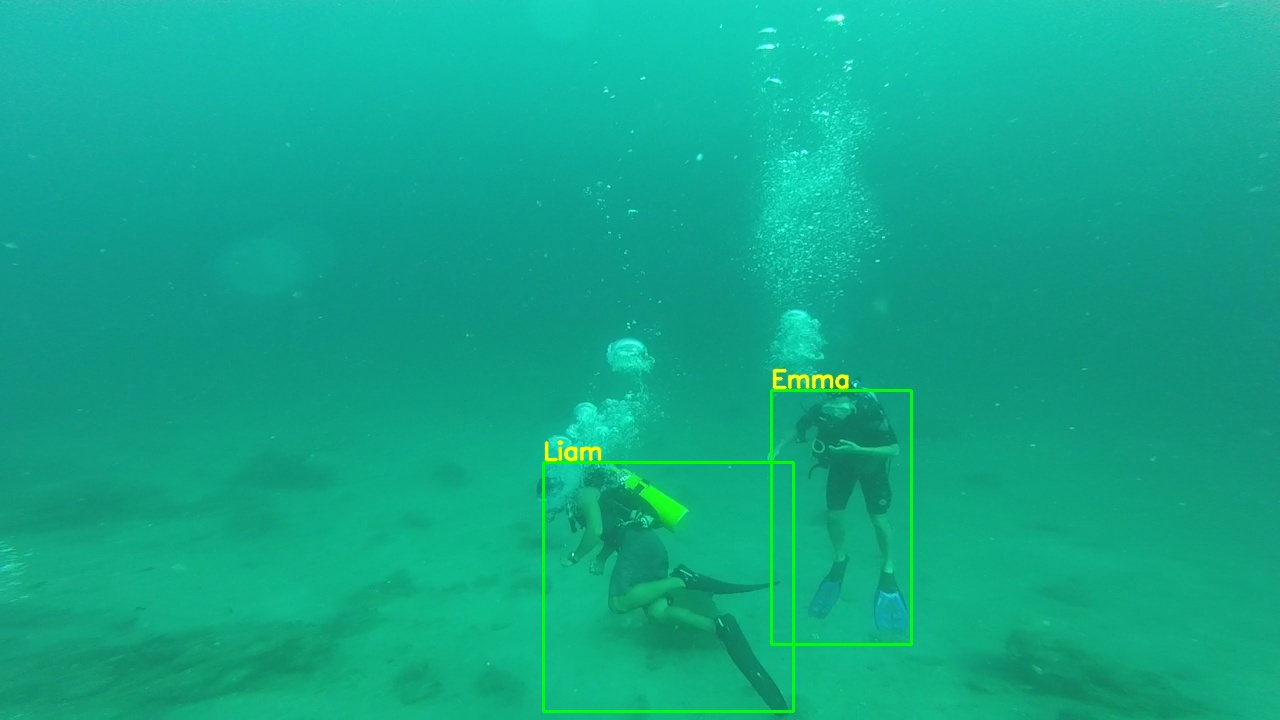}
	\end{subfigure} 
	\caption{Detecting divers Emma and Liam in SCUBA gear in another ocean setting with different color and scale characteristics.}
	\label{fig:twoDiversOceanTwo}
\end{figure}

Overall, the proposed algorithm is found to be highly accurate in identifying divers in different water conditions. Figures~\ref{fig:twoDiversNoFlippersResume} to~\ref{fig:twoDiversOceanTwo} show qualitative results of the diver identification process. Additionally, Table~\ref{table:performanceTable} compares the accuracy of the proposed approach across eight scenarios. Correct identification is above $90$\% for six of the eight scenarios. The worst accuracy is $77.5$\% when tracking three divers with one leaving the scene (scenario $5$). Other than this scenario and scenario $6$, the identification accuracy is high, which makes the approach feasible for underwater human-robot collaborative applications. There are two possible reasons why scenarios $5$ and $6$ have low accuracy: first, bubbles produced by three swimmers (a larger volume than from bubbles produced by two swimmers) obstruct the visual detection of features of each swimmer, which may lead to reduced detection accuracy. Second, since the three swimmers are very close, bounding boxes of swimmers drawn during the general diver detection stage can sometimes overlap which may dilute the difference of features extracted from swimmers.

\subsection{Training and Inference Performance}
We trained the Faster R-CNN detector on a quad-GPU (NVIDIA 1080) system for $200,000$ iterations, which required $10$ hours. The algorithm achieves a run-time of $2.414$ FPS on an Intel Core i7-$5930$K CPU running at $3.50$ GHz. For applications on an AUV, this is acceptable performance, though we achieve further performance improvements via code optimization and a C++ implementation. 

We have also attached a video showing our method in action on sequences of multiple divers.

\section{Conclusions}
\label{sec:conclusions} 
This paper presents an approach for uniquely identifying divers in visual scenes using a combination of feature-based and deep convolutional detection models. A fast and reliable deep detection model is first used to find regions in an image containing divers. Once obtained, a set of spatial and frequency-domain features are then extracted from each of these regions to uniquely identify the diver contained therein. We demonstrate the accuracy of the algorithm on handcrafted experimental scenarios in closed-water environments and also show that it is able to identify divers in both open-water and closed-water environments and under varying diver appearances. 

While this work proposes the first vision-based algorithm to uniquely identify divers, it is also part of a larger framework for human-robot communication, enabling AUVs to interact only with the particular users allowed to instruct the robot. To that effect, future work will integrate gesture-based communication and diver-following abilities with the diver-identification features. We are also currently working on enhancing the accuracy of the deep diver detection models while requiring less computational resources for robot deployment in open-water trials. 
\section*{Acknowledgement}
\label{sec:acknowledgement} 
We gratefully acknowledge the support of NVIDIA Corporation with the donation of the Titan Xp GPU used for this research and the support of the MnDrive initiative. We also acknowledge colleagues Marc Ho, Julian Lagman, and Hannah Dubois for assisting with pool trials and providing test datasets.
%
\clearpage
\bibliographystyle{plain}
\bibliography{allbibs}

\begin{thebibliography}{10}

\bibitem{bao2005canny}
Paul Bao, Lei Zhang, and Xiaolin Wu.
\newblock Canny edge detection enhancement by scale multiplication.
\newblock {\em IEEE transactions on pattern analysis and machine intelligence},
  27(9):1485--1490, 2005.

\bibitem{bay2006surf}
Herbert Bay, Tinne Tuytelaars, and Luc Van~Gool.
\newblock {SURF: Speeded Up Robust Features}.
\newblock {\em European Conference on Computer Vision {ECCV 2006}}, pages
  404--417, 2006.

\bibitem{berg2008computational}
Mark~de Berg, Otfried Cheong, Marc~van Kreveld, and Mark Overmars.
\newblock {\em {Computational Geometry: Algorithms and Applications}}.
\newblock Springer-Verlag TELOS, 2008.

\bibitem{chen2017implementation}
Xinlei Chen and Abhinav Gupta.
\newblock {An implementation of Faster R-CNN with study for region sampling}.
\newblock {\em arXiv preprint arXiv:1702.02138}, 2017.

\bibitem{cormen2009introduction}
Thomas~H Cormen, Charles~E Leiserson, Ronald~L Rivest, and Clifford Stein.
\newblock {\em {Introduction to Algorithms}}.
\newblock MIT press, 2009.

\bibitem{dayoub2015robotic}
Feras Dayoub, Matthew Dunbabin, and Peter Corke.
\newblock {Robotic detection and tracking of Crown-of-Thorns Starfish}.
\newblock In {\em Proceedings of the IEEE/RSJ International Conference on
  Intelligent Robots and Systems (IROS)}, pages 1921--1928. IEEE, 2015.

\bibitem{douglas1973algorithms}
David~H Douglas and Thomas~K Peucker.
\newblock Algorithms for the reduction of the number of points required to
  represent a digitized line or its caricature.
\newblock {\em Cartographica: The International Journal for Geographic
  Information and Geovisualization}, 10(2):112--122, 1973.

\bibitem{Fabbri2018ICRA}
Cameron Fabbri, Md~Jahidul Islam, and Junaed Sattar.
\newblock {Enhancing Underwater Imagery using Generative Adversarial Networks}.
\newblock In {\em Proceedings of the {IEEE International Conference on Robotics
  and Automation (ICRA)}, to appear}, Brisbane, Queensland, Australia, May
  2018.

\bibitem{he2018mask}
Kaiming He, Georgia Gkioxari, Piotr Doll{\'a}r, and Ross Girshick.
\newblock {Mask R-CNN}.
\newblock {\em IEEE transactions on Pattern Analysis and Machine Intelligence},
  2018.

\bibitem{hu1962visual}
Ming-Kuei Hu.
\newblock {Visual Pattern Recognition by Moment Invariants}.
\newblock {\em IRE Transactions on Information Theory}, 8(2):179--187, 1962.

\bibitem{hutchinson96tutorial}
S.~A. Hutchinson, G.~D. Hager, and P.~I. Corke.
\newblock A tutorial on visual servo control.
\newblock {\em IEEE Transactions on Robotics and Automation}, 12(5):651--670,
  10 1996.

\bibitem{Isard98Condensation}
Michael Isard and Andrew Blake.
\newblock {CONDENSATION} -- conditional density propagation for visual
  tracking.
\newblock {\em International Journal of Computer Vision}, 29(1):5--28, 1998.

\bibitem{islam2018Hands}
Md~Jahidul Islam, Marc Ho, and Junaed Sattar.
\newblock Dynamic reconfiguration of mission parameters in underwater
  human-robot collaboration.
\newblock In {\em Proceedings of the 2018 IEEE International Conference on
  Robotics and Automation (ICRA), to appear.} IEEE, 2018.

\bibitem{islam2018person}
Md~Jahidul Islam, Jungseok Hong, and Junaed Sattar.
\newblock Person following by autonomous robots: A categorical overview.
\newblock {\em arXiv preprint arXiv:1803.08202}, 2018.

\bibitem{islam2017mixed}
Md~Jahidul Islam and Junaed Sattar.
\newblock Mixed-domain biological motion tracking for underwater human-robot
  interaction.
\newblock In {\em Proceedings of the 2017 IEEE International Conference on
  Robotics and Automation (ICRA)}, pages 4457--4464. IEEE, 2017.

\bibitem{julier97new}
Simon Julier and Jeffrey~K. Uhlmann.
\newblock {A new extension of the Kalman filter to nonlinear systems}.
\newblock {\em Signal processing, sensor fusion, and target recognition VI},
  pages 182--193, 1997.

\bibitem{Kalman60}
Rudolph~Emil Kalman.
\newblock A new approach to linear filtering and prediction problems.
\newblock {\em Transactions of the {ASME--Journal of Basic Engineering}},
  82(Series D):35--45, 1960.

\bibitem{liu2016ssd}
Wei Liu, Dragomir Anguelov, Dumitru Erhan, Christian Szegedy, Scott Reed,
  Cheng-Yang Fu, and Alexander~C Berg.
\newblock {SSD: Single shot multibox detector}.
\newblock In {\em European conference on computer vision}, pages 21--37.
  Springer, 2016.

\bibitem{lloyd1982least}
Stuart Lloyd.
\newblock {Least squares quantization in PCM}.
\newblock {\em IEEE transactions on information theory}, 28(2):129--137, 1982.

\bibitem{nixon_human_2005}
M.~S. Nixon, T.~N. Tan, and R.~Chellappa.
\newblock {\em Human Identification Based on Gait}.
\newblock The Kluwer International Series on Biometrics. Springer-Verlag New
  York, Inc. Secaucus, NJ, USA, 2005.

\bibitem{niyogi_analyzing_1994}
S.~A. Niyogi and E.~H. Adelson.
\newblock {Analyzing and recognizing walking figures in XYT}.
\newblock In {\em Proceedings of IEEE Computer Society Conference on Computer
  Vision and Pattern Recognition}, pages 469--474, 1994.

\bibitem{Rashid1980}
R.F. Rashid.
\newblock Toward a system for the interpretation of moving light display.
\newblock {\em IEEE Transactions on Pattern Analysis and Machine Intelligence},
  2(6):574--581, November 1980.

\bibitem{redmon2016yolo9000}
Joseph Redmon and Ali Farhadi.
\newblock {YOLO9000: Better, Faster, Stronger}.
\newblock {\em arXiv preprint arXiv:1612.08242}, 2016.

\bibitem{tinyYOLO}
Joseph Redmon and Ali Farhadi.
\newblock {Tiny YOLO}.
\newblock \url{https://pjreddie.com/darknet/yolo/}, 2017.
\newblock Accessed: 2-20-2018.

\bibitem{NIPS2015_5638}
Shaoqing Ren, Kaiming He, Ross Girshick, and Jian Sun.
\newblock {Faster R-CNN: Towards Real-Time Object Detection with Region
  Proposal Networks}.
\newblock In C.~Cortes, N.~D. Lawrence, D.~D. Lee, M.~Sugiyama, and R.~Garnett,
  editors, {\em Advances in Neural Information Processing Systems 28}, pages
  91--99. Curran Associates, Inc., 2015.

\bibitem{rublee2011orb}
Ethan Rublee, Vincent Rabaud, Kurt Konolige, and Gary Bradski.
\newblock {ORB: an efficient alternative to SIFT or SURF}.
\newblock In {\em Computer Vision (ICCV), 2011 IEEE International Conference
  on}, pages 2564--2571. IEEE, 2011.

\bibitem{Sattar07IROS}
Junaed Sattar and Gregory Dudek.
\newblock Where is your dive buddy: tracking humans underwater using
  spatio-temporal features.
\newblock In {\em Proceedings of the {IEEE/RSJ} International Conference on
  Intelligent Robots and Systems ({IROS})}, pages 3654--3659, San Diego,
  California, USA, October 2007.

\bibitem{Sattar09RSS}
Junaed Sattar and Gregory Dudek.
\newblock Underwater human-robot interaction via biological motion
  identification.
\newblock In {\em Proceedings of the International Conference on Robotics:
  Science and Systems V, {RSS}}, pages 185--192, Seattle, Washington, USA, June
  2009. MIT Press.

\bibitem{Sattar05IROS}
Junaed Sattar, Philippe Gigu\`ere, Gregory Dudek, and Chris Prahacs.
\newblock A visual servoing system for an aquatic swimming robot.
\newblock In {\em Proceedings of the IEEE/RSJ International Conference on
  Intelligent Robots and Systems ({IROS})}, pages 1483--1488, Edmonton,
  Alberta, Canada, 8 2005.

\bibitem{Sattar14ICRA}
Junaed Sattar and James~Joseph Little.
\newblock {Ensuring Safety in Human-Robot Dialog -- a Cost-Directed Approach}.
\newblock In {\em Proceedings of the IEEE International Conference on Robotics
  and Automation, {ICRA}.}, pages 6660--6666, Hong Kong, China, May 2014.

\bibitem{shkurti2017underwater}
F.~Shkurti, W.~D. Chang, P.~Henderson, M.~J. Islam, J.~C.~G. Higuera, J.~Li,
  T.~Manderson, A.~Xu, G.~Dudek, and J.~Sattar.
\newblock Underwater multi-robot convoying using visual tracking by detection.
\newblock In {\em 2017 IEEE/RSJ International Conference on Intelligent Robots
  and Systems (IROS)}, pages 4189--4196, Sept 2017.

\bibitem{Sidenbladh2003}
Hedvig Sidenbladh and Michael~J. Black.
\newblock Learning the statistics of people in images and video.
\newblock {\em International Journal of Computer Vision}, 54(1-3):181--207,
  2003.

\bibitem{Sidenbladh00Stochastic}
Hedvig Sidenbladh, Michael~J. Black, and David~J. Fleet.
\newblock {Stochastic tracking of 3D human figures using 2D image motion}.
\newblock In {\em Proceedings of the European Conference on Computer Vision},
  volume~2, pages 702--718, 2000.

\bibitem{tfzoo}
Tensorflow.
\newblock Tensorflow object detection zoo.
\newblock
  \url{https://github.com/tensorflow/models/blob/master/research/object_detection/g3doc/detection_model_zoo.md},
  2017.
\newblock Accessed: 2-20-2018.

\bibitem{Sattar08ICRA}
Anqi Xu, Gregory Dudek, and Junaed Sattar.
\newblock A natural gesture interface for operating robotic systems.
\newblock In {\em Proceedings of the IEEE International Conference on Robotics
  and Automation, {ICRA}}, pages 3557--3563, Pasadena, California, May 2008.

\end{thebibliography}
\end{document}